\documentclass[conf]{new-aiaa}

\usepackage{graphicx}
\usepackage{amsmath}
\usepackage[version=4]{mhchem}
\usepackage{siunitx}
\usepackage{longtable,tabularx}
\usepackage{todonotes}
\usepackage{subcaption}

\newcommand{\FIG}[1] {Figure~\ref{#1}}
\newcommand{\SEC}[1] {Section~\ref{#1}}

\usepackage{caption}

\title{Distributed State Estimation for Vision-based Cooperative Slung Load Transportation in GPS-denied Environments}

\author{Jack Pence,\footnote{Graduate Student, Department of Aerospace Engineering, Student Member AIAA}
    Jackson Fezell,\footnote{Graduate Student, Department of Aerospace Engineering, Student Member AIAA}
    Jack W. Langelaan,\footnote{Professor of Aerospace Engineering, Associate Fellow AIAA}
    and Junyi Geng\footnote{Assistant Professor of Aerospace Engineering, Member AIAA}}
\affil{The Pennsylvania State University, University Park, Pennsylvania, 16802}

\begin{document}

\begin{center}
{\Large\textcolor{red}{AIAA SciTech Forum, January 12-16, 2026}\\
\textcolor{purple}{\url{https://arc.aiaa.org/doi/abs/10.2514/6.2026-2575}}}
\end{center}

\maketitle

\begin{abstract}
Transporting heavy or oversized slung loads using rotorcraft has traditionally relied on single-aircraft systems, which limits both payload capacity and control authority. Cooperative multilift using teams of rotorcraft offers a scalable and efficient alternative, especially for infrequent but challenging “long-tail” payloads without the need of building larger and larger rotorcraft. Most prior multilift research assumes GPS availability, uses centralized estimation architectures, or relies on controlled laboratory motion-capture setups. As a result, these methods lack robustness to sensor loss and are not viable in GPS-denied or operationally constrained environments. This paper addresses this limitation by presenting a distributed and decentralized payload state estimation framework for vision-based multilift operations. Using onboard monocular cameras, each UAV detects a fiducial marker on the payload and estimates its relative pose. These measurements are fused via a Distributed and Decentralized Extended Information Filter (DDEIF), enabling robust and scalable estimation that is resilient to individual sensor dropouts. This payload state estimate is then used for closed-loop trajectory tracking control. Monte Carlo simulation results in Gazebo show the effectiveness of the proposed approach, including the effect of communication loss during flight.
\end{abstract}

\section{Introduction} \label{s:intro}


The transport of externally slung loads has been a key capability of rotorcraft for many decades~\cite{superpumafirefighting, hutto1976flight}. This method of cargo transport addresses a variety of challenges, such as payloads being too large to fit into the cargo bay, rotorcraft unable to land at the delivery site, or the need to operate as a ground-based crane. These missions require the use of a pilot, an expensive vehicle, and careful coordination to complete. If it is a combat mission, then the risk only increases, and payload transport by an autonomous system could help safeguard human life.

Traditionally, slung load transportation has been accomplished using a single rotorcraft with one or more cables attached to a payload. This approach limits both payload controllability and the maximum weight that can be carried~\cite{murray1996trajectory}. Heavy, ``long tail'' payloads, such as those handled by the Russian Mi-26 helicopter, cannot be lifted by standard rotorcraft~\cite{smirnov1990multiple}. While specialized larger and higher lifting capacity rotorcraft could be developed to solve this problem, it is not economic to develop such large vehicles for these infrequent payloads. Furthermore, it has been shown that as the rotorcraft size increases, the gain in relative productivity diminishes, making oversized vehicles an inefficient solution for ``long tail'' payload transportation~\cite{carter1982implication}.

\begin{figure}
\vspace{-2.5mm}
  \centering  
    \includegraphics[width=\textwidth]{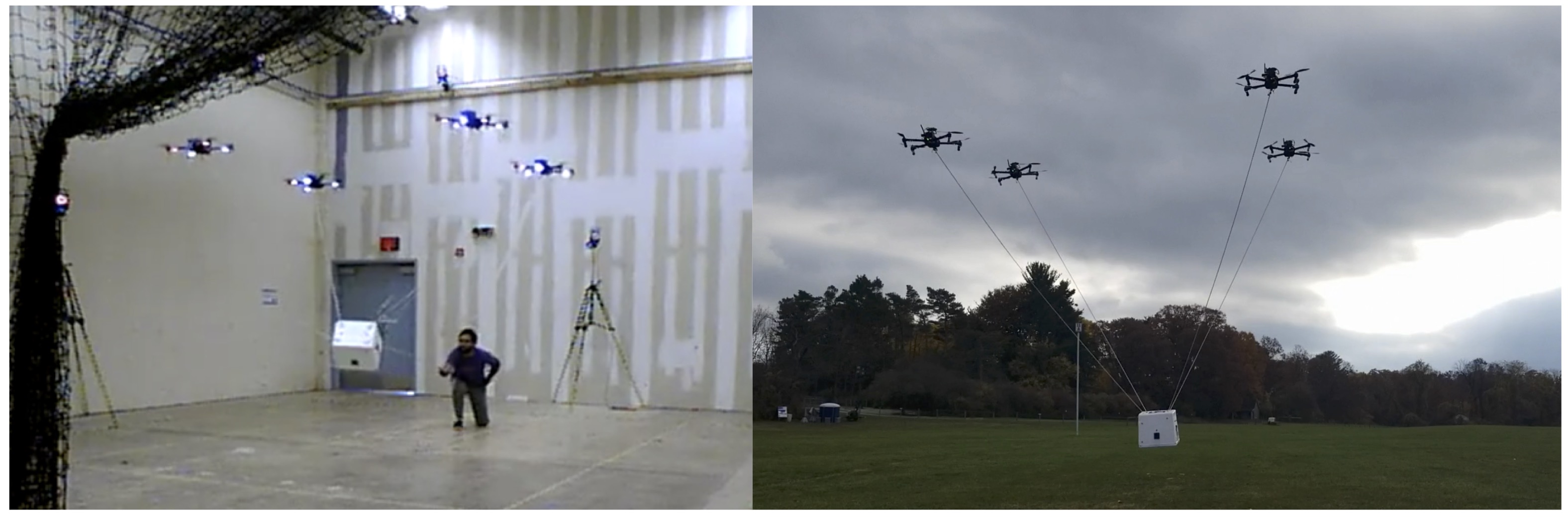}
  \caption{Left: multilift system showing disturbance rejection; right: outdoor demonstration of multilift~\cite{geng2020cooperative,  GenLan19, geng2021estimation, geng2022load}} 
  \label{fig:psu_multilift}
\vspace{-1.5mm}
\end{figure}

Another way to increase payload capacity is transporting the slung load using a team of cooperating rotorcraft (\FIG{fig:psu_multilift}). This concept of multilift enables smaller rotorcraft (developed to carry more common payloads) to work together to transport heavier, ``long tail'' payloads, providing a more economical and efficient solution. In addition to increased payload capacity, such systems have several other advantages including mission redundancy, robustness, and overall operational performance.

There has been a significant amount of work over many decades examining multilift, which began in 1970 using a pair of CH-54B helicopters in the twin lift configuration~\cite{MeeChe70}. Subsequent work focused on dynamics of the twin lift system and the use of spreader bars to increase payload capacity and controllability~\cite{CicKan92}, ~\cite{MitPra91,MitPra92,MitPra93}. In 2009, Kondak et al. described the use of more than two helicopters to carry a single payload~\cite{BerKon09}.

More recent work has focused on formation control by treating either the helicopter or the payload as a point mass~\cite{BerKon09, MicFin11}, controlling the full payload vehicle dynamics with ``load-leading'' control~\cite{LiHor14, GenLan19, geng2020cooperative}, and operating under parameter uncertainty~\cite{sanalitro2020full, geng2021estimation} or under limited communication ability~\cite{zhang2021self}. However, most multilift approaches have been explored only in static environments using GPS, simulation, or controlled laboratory setups with indoor motion capture. Practical state estimation was not taken into account, and as a result, few of these implementations can be extended easily in the GPS-denied environment.

Although the use of computer vision for UAS navigation as well as relative positioning has been demonstrated by many researchers, (examples include vision-based simultaneous localization and mapping~\cite{zhao2021super, qin2018vins}, exploration~\cite{hu2023off}, autonomous shipboard landing~\cite{horn2023experimental, nicholson2022scaled, hendrick2022scaled}, and manufacturing~\cite{liu2022robotic, deng2022icaps}), the research activity in vision for practical multilift has just recently begun~\cite{li2021cooperative}. However, the early work relies on centralized state estimation for payload states, and will be fragile to sensor data loss from even one UAV.

This paper proposes a distributed and decentralized payload state estimator based on the extended information filter (DDEIF)~\cite{Mutambara1998}. Rather than propagating an estimate of the state and its associated covariance like the commonly used Kalman filter, the DDEIF propagates the information vector and the information matrix. This adds some complications (for example, one must recover the state from the information vector for control) but has significant benefits in decentralization. Because the filter only needs to take care of the information from each of the individual UAVs, it will be robust even if there is sensor data dropout. A similar approach is the Unscented Information Filter (UIF), described in~\cite{Lee08}.

Here, the DDEIF is used to estimate the payload state in vision-based multilift. A fiducial marker (an AprilTag~\cite{OlsonAprilTag2011}) is placed on the payload; a camera and vision processing algorithm on board each UAV then provides the pose of the AprilTag (and hence payload) in the camera frame of that UAV. Measurements from each UAV in the multilift formation are fused in the DDEIF and the resulting payload state estimate can be used for payload control.


The remainder of this paper is organized as follows: \SEC{s:problem} defines coordinate frames, sensor models, and system dynamics; \SEC{s:info_filter} describes the information filter in the context of multilift; \SEC{s:simulation} gives results of simulations conducted in Gazebo; finally \SEC{s:conclusion} presents concluding remarks.

\section{The problem of distributed estimation} \label{s:problem}

In prior multilift implementations~\cite{GenLan19, geng2020cooperative}, the payload pose was assumed to be known, obtained from an onboard sensor suite consisting of an IMU and GPS mounted on the payload, or from a motion capture system. While this configuration is feasible in practice, it introduces several limitations, including the need to physically mount a sensor suite on the payload, communicate pose data to the rotorcraft, and rely on external positioning infrastructure such as GPS. In this work, the payload state is estimated using only a monocular camera on each quadrotor and a single AprilTag on the payload, offering a simpler and more practical configuration that enables robust operation in GPS-denied environments.

\subsection{Multilift control architecture}

\begin{figure}
\vspace{-2.5mm}
  \centering  
    \includegraphics[width=0.9\textwidth]{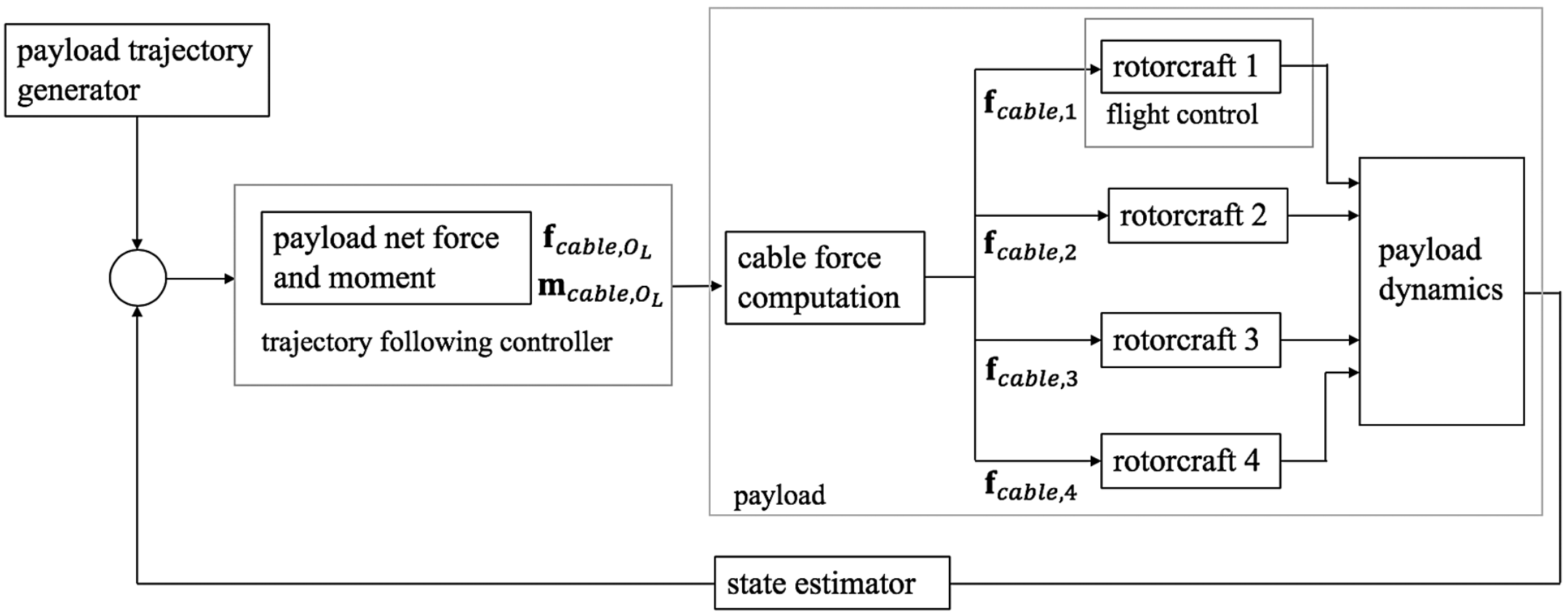}
  \caption{Multilift system block diagram~\cite{geng2020cooperative}.} 
  \label{fig:blockDiag}
\vspace{-1.5mm}
\end{figure}

A distinguishing feature of the multilift control approach is its load-leading design: instead of commanding flight paths for the rotorcraft and allowing the payload to swing passively, the controller commands a payload trajectory, and the rotorcraft act as actuators to realize the desired payload motion. The overall control diagram is illustrated in Figure~\ref{fig:blockDiag}. The work in this paper focuses on replacing the payload state feedback from the onboard sensor suite with a vision-based distributed and decentralized estimation approach, while maintaining the existing control design.

\subsection{Vision-based state estimation scenario}
\begin{figure}
\vspace{-2.5mm}
  \centering  
    \includegraphics[width=0.6\textwidth]{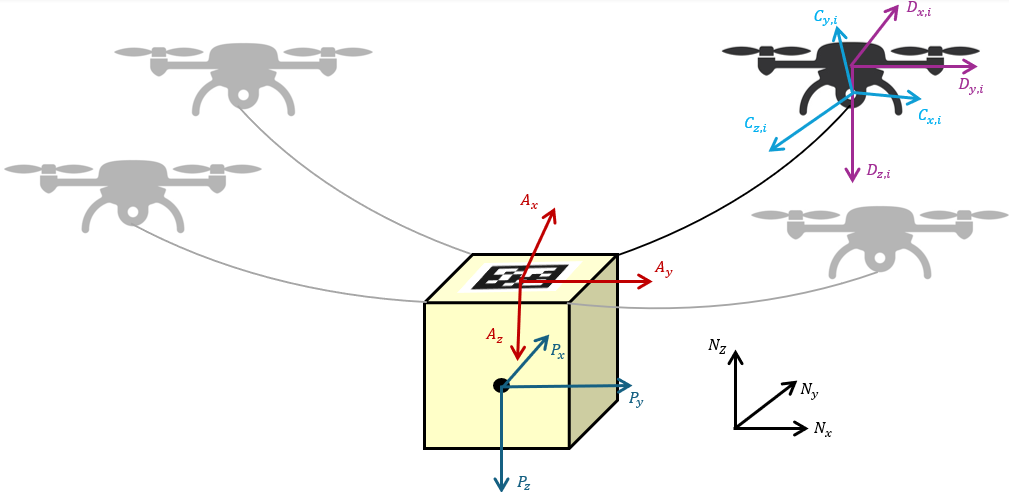}
  \caption{Multilift coordinate frame schematic.} 
  \label{fig:coordinateFrames}
\vspace{-1.5mm}
\end{figure}

In the proposed multilift configuration, each rotorcraft is equipped with a payload-facing monocular camera, and the payload is marked with an AprilTag fiducial marker~\cite{OlsonAprilTag2011}. As the vehicles maintain formation during flight, the cameras continuously observe the AprilTag, providing relative pose measurements between each rotorcraft and the payload. These relative measurements are then fused across the rotorcraft to estimate the global state of the payload in real time. This scenario enables cooperative state estimation without relying on external positioning systems or additional onboard sensor suites. A schematic of the AprilTag configuration within the multilift system is shown in Figure~\ref{fig:coordinateFrames}.

\subsection{Coordinate frames}
In this estimation scenario, accurate transformations between multiple coordinate frames are essential for fusing onboard sensor data into a consistent payload state estimate. These transformations are carried out through a sequence of rotation matrices and known geometric offsets derived from the system configuration. A schematic of the relevant coordinate frame can be found in Figure~\ref{fig:coordinateFrames}.

The primary coordinate frames used in this system are defined as follows: the quadrotor body frame, $D$, is centered at the quadrotor center of gravity with the axes defined as forwards, right, down; the camera frame, $C$, is rigidly attached, vertically offset from the quadrotor center of gravity, and tilted down relative to the rotorcraft body frame with axes defined as right, down, out of lens; the inertial frame, $N$, is centered at the simulation origin with axes defined following the aerospace convention: North, East, and Down (NED).

To transform a vector from the body frame, $D$, to the inertial frame, $N$, the following operation is used:
\begin{equation}
    \mathbf{r}_N = \mathbf{R}_{ND} \mathbf{r}_D
\end{equation}
\begin{equation}
    \mathbf{R}_{ND} =
    \begin{bmatrix}
        \cos\theta \cos\psi & \cos\theta \sin\psi & -\sin\theta \\
        \sin\phi \sin\theta \cos\psi - \cos\phi \sin\psi & \sin\phi \sin\theta \sin\psi + \cos\phi \cos\psi & \sin\phi \cos\theta \\
        \cos\phi \sin\theta \cos\psi + \sin\phi \sin\psi & \cos\phi \sin\theta \sin\psi - \sin\phi \cos\psi & \cos\phi \cos\theta
    \end{bmatrix}
\end{equation}
where $\mathbf{R}_{ND}$ is the rotation matrix representing the rotorcraft’s orientation relative to the inertial frame and $\phi$, $\theta$, and $\psi$ correspond to the roll, pitch, and yaw of the rotorcraft relative to the inertial frame.

The transformation from the camera frame to the body frame incorporates both the camera's fixed tilt angle, $\alpha$, and the camera frame coordinate system: right, down, and out of lens. The corresponding rotation from the camera frame, $C$, to the body frame, $D$, is:
\begin{equation}
    \mathbf{R}_{DC} =
    \begin{bmatrix}
        0 & -\cos\alpha & \sin\alpha \\
        1 & 0 & 0 \\
        0 & \sin\alpha & \cos\alpha
    \end{bmatrix}
\end{equation}

\subsection{Sensor models}

To obtain the relative pose of the AprilTag marker with respect to the camera frame, the AprilTag 3 detection algorithm is employed~\cite{OlsonAprilTag2011}. Given an input image of the tag along with the camera specifications and the tag's physical dimensions, the algorithm computes the tag’s position and orientation (pose) relative to the camera. The global pose of the tag can then be reconstructed by combining these relative measurements with the quadrotor’s pose in the inertial frame. This transformation accounts for the camera’s rotation and translation relative to the quadrotor body, the quadrotor’s pose in the inertial frame, and the fixed offset between the AprilTag and the payload’s center of gravity, $\mathbf{s}_p$.

The payload position measurement, $\mathbf{z}_p$, relative to the inertial origin, is obtained by combining the AprilTag position measurement with the quadrotor’s position in the inertial frame. Accurate knowledge of the quadrotor position is required for this calculation, which can be obtained using mature visual-inertial odometry (VIO) algorithms such as VINS-Mono, VINS-Fusion~\cite{qin2018vins}, or SuperOdometry~\cite{zhao2021super}. The implementation of these algorithms is outside the scope of this paper.

\begin{figure}
\vspace{-2.5mm}
  \centering  
    \includegraphics[width=0.6\textwidth]{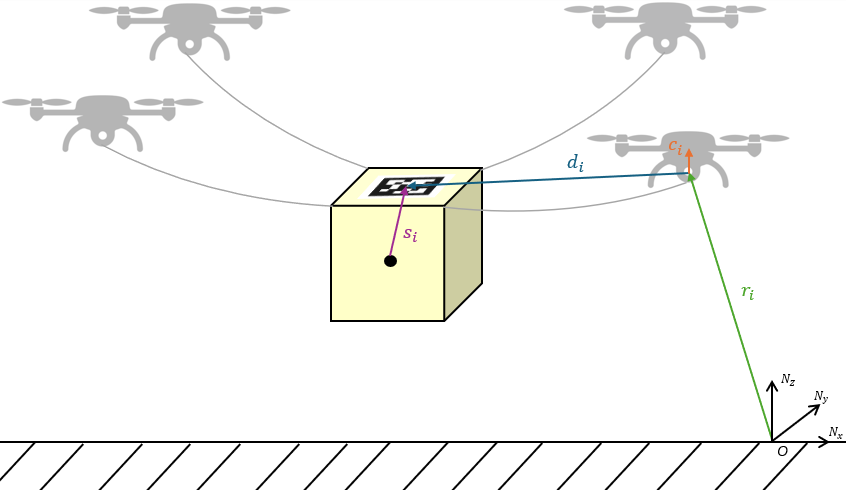}
  \caption{Multilift payload measurement calculation vectors.} 
  \label{fig:measurementCalculation}
\vspace{-1.5mm}
\end{figure}

To compute the position measurement, the following sequence of transformations is applied:
\begin{equation}
    \mathbf{z}_p = \mathbf{R}_{ND} (\mathbf{R}_{DC} \mathbf{d}_C + \mathbf{c}_D) + \mathbf{r}_N
\end{equation}
where $\mathbf{d}_C$ is the AprilTag measurement in the camera frame, $\mathbf{c}_D$ is the displacement of the camera from the quadrotor center of gravity, $\mathbf{R}_{DC}$ is the rotation from the camera frame to the body frame, $\mathbf{R}_{ND}$ is the rotation from body frame to inertial frame, and $\mathbf{r}_N$ is the position of the quadrotor in the inertial frame. The fixed offset between the AprilTag and the payload center of gravity, $\mathbf{s}_P$, is applied after the AprilTag state is transformed. A schematic of these vectors and transformations is shown in Figure~\ref{fig:measurementCalculation}.

The orientation measurement, $\mathbf{z}_q$, relative to the inertial frame, is obtained by transforming the relative AprilTag orientation from the camera frame. This transformation accounts for the fixed tilt of the camera relative to the quadrotor body frame and the quadrotor's orientation relative to the inertial frame which can be obtained using the same VIO algorithms discussed above. The position and orientation measurements are then concatenated to form the complete payload measurement, $\mathbf{z} = [\mathbf{z}_p, \mathbf{z}_q]$, which is used in the state estimator described in Section~\ref{s:info_filter}.

\subsection{Payload model}
The payload controller requires the payload’s position, orientation, velocity, and angular rates to compute its control input. Therefore, these quantities are selected as the estimator’s state variables:
\begin{equation}
\label{system:state}
    \mathbf{x} = [n,e,d,q_w,q_x,q_y,q_z,v_n,v_e,v_d,P,Q,R]^\top
\end{equation}

In the multilift load-leading control architecture, the payload control inputs consist of its linear and angular accelerations:
\begin{equation}
\label{system:input}
    \mathbf{u} = [\dot{v}_n,\dot{v}_e,\dot{v}_d,\dot{P},\dot{Q},\dot{R}]^\top
\end{equation}

Utilizing the following payload kinematics and dynamics, the state estimator's process model is defined as:
\begin{equation}
\label{state_dynamics:position}
    [\dot{n},\dot{e},\dot{d}]^\top = [v_n,v_e,v_d]^\top
\end{equation}
\begin{equation}
\label{state_dynamics:quaternion}
\begin{bmatrix}
\dot{q}_w \\ \dot{q}_x \\ \dot{q}_y \\ \dot{q}_z 
\end{bmatrix}
= \frac{1}{2}
\begin{bmatrix}
0 & -P & -Q & -R \\
P & 0 & R & -Q \\
Q & -R & 0 & P \\
R & Q & -P & 0 
\end{bmatrix}
\begin{bmatrix}
q_w \\ q_x \\ q_y \\ q_z 
\end{bmatrix}
\end{equation}
\begin{equation}
\label{state_dynamics:velocity}
    [\dot{v}_n,\dot{v}_e,\dot{v}_d]^\top = [\dot{v}_n,\dot{v}_e,\dot{v}_d]^\top_{\mathbf{u}}
\end{equation}
\begin{equation}
\label{state_dynamics:omega}
    [\dot{P},\dot{Q},\dot{R}]^\top = [\dot{P},\dot{Q},\dot{R}]^\top_{\mathbf{u}}
\end{equation}

\section{Distributed and Decentralized Extended Information Filter} \label{s:info_filter}
To fuse the payload measurements from multiple quadrotors, a distributed and decentralized extended information filter is employed. This approach is well suited to the multilift scenario because it allows each quadrotor to maintain its own state estimate while incorporating measurements from the other vehicles, without relying on a central or lead quadrotor. The distributed structure enhances robustness to communication delays or dropouts, avoids dependence on a central node, and scales naturally if additional quadrotors were added to the system.

A more detailed derivation of the information filter can be found in~\cite{Mutambara1998}, but most simply put, an information filter can be thought of as an inverse Kalman filter. Rather than propagating the payload state, $\mathbf{x}$, and a covariance, $\mathbf{P}$, through time, an information filter propagates an information matrix, $\mathbf{Y}$, and an associated information vector, $\mathbf{y}$:
\begin{equation}
\label{information:transformation}
    \mathbf{Y} = \mathbf{P}^{-1}
\end{equation}
\begin{equation}
\label{information:transformation2}
    \mathbf{y} = \mathbf{P}^{-1}\mathbf{x}
\end{equation}
At each time step, $k$, the information matrix and information vector of the payload are propagated on each quadrotor with:
\begin{equation}
\label{information:prediction}
    \mathbf{y}_{k|k-1} = \mathbf{Y}_{k|k-1} \mathbf{f}(k,\mathbf{x}_{k-1|k-1},\mathbf{u}_{k-1})
\end{equation}
\begin{equation}
\label{information:prediction2}
    \mathbf{Y}_{k|k-1} = [\nabla{\mathbf{f_x}_{,k}} \mathbf{Y}^{-1}_{k-1|k-1} \nabla{\mathbf{f_x}_{,k}}^\top + \mathbf{Q}_k]^{-1}
\end{equation}
where $\mathbf{f}_k$ is the nonlinear state transition function for the payload defined by Eq.~\ref{state_dynamics:position}-~\ref{state_dynamics:omega}, and $\mathbf{Q}_k$ is the process noise covariance matrix.

Each quadrotor in the multilift system maintains its own information vector and information matrix, $\mathbf{y}_{k|k}$ and $\mathbf{Y}_{k|k}$. When available, each new measurement is used to calculate a measurement information vector, $\mathbf{i}_k$, and a measurement information matrix, $\mathbf{I}_k$ on the respective quadrotor:
\begin{equation}
\label{information:I}
    \mathbf{I}_{k} = \nabla{\mathbf{h_x}_{,k}}^\top \mathbf{R}^{-1}_{k} \nabla{\mathbf{h_x}_{,k}}
\end{equation}
\begin{equation}
\label{information:i}
    \mathbf{i}_{k} = \nabla{\mathbf{h_x}_{,k}}^\top \mathbf{R}^{-1}_{k} [\mathbf{\nu}_k + \nabla{\mathbf{h_x}_{,k}} \mathbf{x}_{k|k-1}]
\end{equation}
\begin{equation}
\label{information:nu}
    \mathbf{\nu}_{k} = \mathbf{z}_k - \mathbf{h}(\mathbf{x}_{k|k-1})
\end{equation}
where $\mathbf{h}_{k}$ is the nonlinear measurement function, $\mathbf{R}_{k}$ is the measurement noise covariance matrix, and $\mathbf{\nu}_{k}$ is the innovation.

In ideal communication conditions, each quadrotor, $j$, shares its calculated measurement information, $\mathbf{i}_k$ and $\mathbf{I}_k$, so that each quadrotor receives and utilizes all available measurement information. However, in some cases when communication between quadrotors is unavailable or intermittent, each quadrotor simply uses its own measurement information and all available measurement information. The total information vector and information matrix are updated using:
\begin{equation}
\label{information:y_update}
    \mathbf{y}_{k|k} = \mathbf{y}_{k|k-1} + \sum_{j=1}^{N}\mathbf{i}_{k,j}
\end{equation}
\begin{equation}
\label{information:Y_update}
    \mathbf{Y}_{k|k} = \mathbf{Y}_{k|k-1} + \sum_{j=1}^{N}\mathbf{I}_{k,j}
\end{equation}
where $N$ is the number of quadrotors with available information. This summation will always include at least a quadrotor's own measurement information as defined by Eq. \ref{information:I} and \ref{information:i}.

\section{Results and discussion} \label{s:simulation}

\begin{figure}
\vspace{-2.5mm}
  \centering  
    \includegraphics[width=0.6\textwidth]{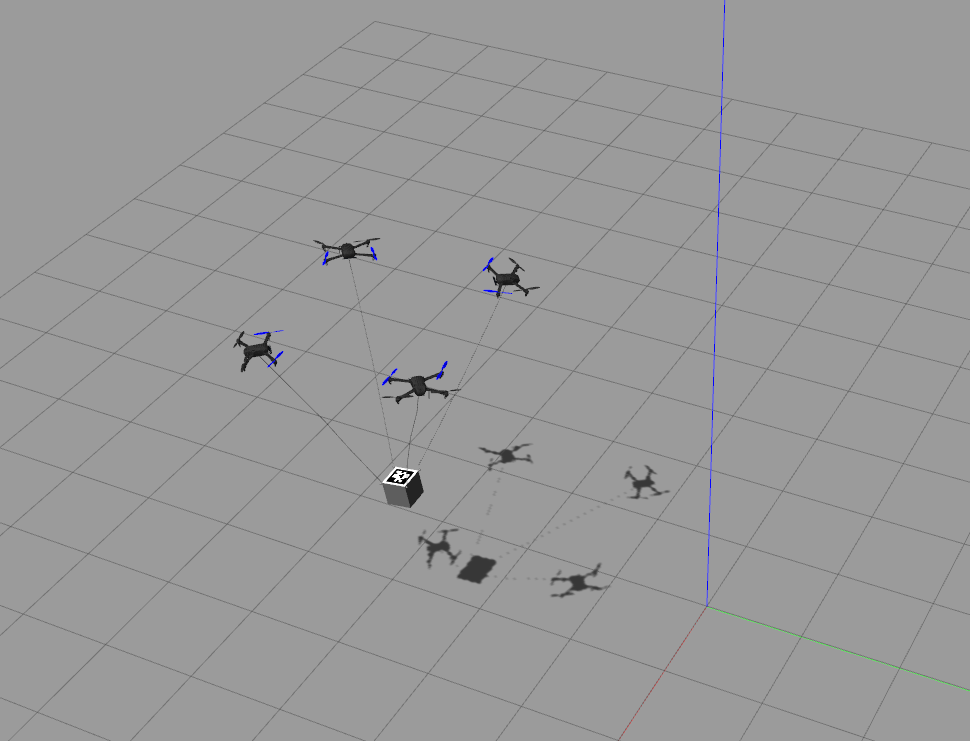}
  \caption{Multilift Gazebo simulation.} 
  \label{fig:gazeboSim}
\vspace{-1.5mm}
\end{figure}

To evaluate the performance of the proposed estimator, software-in-the-loop simulations of the multilift system were conducted. These simulations were performed in Gazebo to provide realistic physics and sensor feedback under high-fidelity conditions. Onboard each rotorcraft, PX4 serves as the autopilot, while ROS is used as the communication framework. A rendering of the multilift system in the Gazebo environment is shown in Figure~\ref{fig:gazeboSim}.

\subsection{Simulation description}
To maintain a realistic simulation environment, a disturbance force is applied at the payload's center of gravity in a randomized direction within the North-East plane. This disturbance is modeled as a Gaussian distribution with a mean of zero and a standard deviation of 0.227 N, chosen to yield small but observable accelerations representative of minor aerodynamic effects. The characteristics of this disturbance were used to compute the process noise covariance matrix, $\mathbf{Q}$, as shown in Eq.~\ref{information:prediction2}. The measurement noise covariance matrix, $\mathbf{R}$, presented in Eq.~\ref{information:I}, was calculated with knowledge of the PX4 rotorcraft pose covariance from the autopilot's onboard extended Kalman filter and an assumed AprilTag pose uncertainty from~\cite{Kalaitzakis2021}. The process noise and measurement noise covariance matrices can be expressed as:
\begin{equation}
\label{covariance:Q}
    \mathbf{Q} = \text{diag}([0.0625, 0.0625, 0.0625, 7.84, 7.84, 7.84])
\end{equation}
\begin{equation}
\label{covariance:R}
    \mathbf{R} = \text{diag}([0.12, 0.12, 0.12, 0.0027, 0.0027, 0.0027])
\end{equation}
noting that the angular terms of the measurement noise covariance here are reported as Euler angle variances (radians squared), and are mapped into the quaternion error covariance used in the DDEIF algorithm.

The parameters used to define the multilift model in Gazebo are summarized in Table~\ref{t:parameters}. These parameters are similar to those used in previous works~\cite{geng2020cooperative} and are provided here for completeness. All simulations were conducted with the multilift control loop and state estimator running at 20 Hz, while the Gazebo physics engine operated at 250 Hz. Monte Carlo simulations were conducted to analyze the performance of the DDEIF, with each simulation consisting of 50 independent runs differing only by the random realization of the payload disturbance. Results from these simulations are presented in the following sections. For readability, only the state estimates from one of the four rotorcraft are shown. All rotorcraft produce similar results since they each run the same estimator software independently.

\begin{table}
\centering
\caption{Parameters for the payload, cables, rotorcraft, cameras, and AprilTag}
\label{t:parameters}
\begin{tabular}{l c}
\hline\hline
\textbf{Parameter} & \textbf{Value} \\
\hline
\multicolumn{2}{l}{\textbf{Payload and Cables}} \\
Payload mass & 1.2 kg \\
Payload size & 0.25 $\times$ 0.25 $\times$ 0.25 m \\
Cable length & 2.0 m \\
\hline
\multicolumn{2}{l}{\textbf{Rotorcraft and Cameras}} \\
Rotorcraft model & 3DR \\
Rotorcraft mass & 1.5 kg \\
Rotorcraft size & 0.27 $\times$ 0.41 $\times$ 0.08 m \\
Camera model & Intel RealSense D455 \\
Camera format & RGB \\
Camera FoV & 1.5 rad \\
Camera Resolution & 848 $\times$ 480 px \\
\hline
\multicolumn{2}{l}{\textbf{AprilTag}} \\
Size & 0.25 m $\times$ 0.25 m \\
Family & 36h11 \\
\hline\hline
\end{tabular}
\end{table}

\subsection{Isolated state estimate performance} \label{s:estimator_performance}
To isolate and evaluate the performance of the proposed state estimator, the multilift controller in this section operates using ground-truth state data provided by the Gazebo simulator, rather than the estimated state. This allows for a focused analysis of estimation accuracy independent of any control loop effects. In Section~\ref{s:estimator_intheloop}, we later evaluate the closed-loop performance when the estimated payload state is used directly in the load-leading controller.

\begin{figure}
  \centering  
    \includegraphics[width=3.5in]{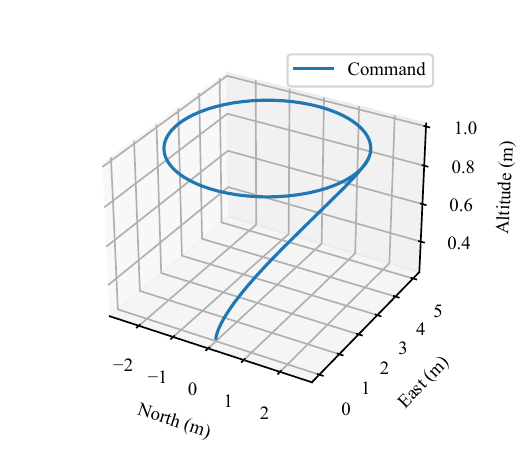}
  \caption{Commanded pirouetting circle trajectory.}
  \label{fig:pirouette_3D_trajectory}
\end{figure}

The trajectory used to evaluate estimator performance is a pirouetting circle in the North-East plane, with a radius of 2.5 meters and a tangential velocity of 0.5 meters per second. One wall of the payload continuously faces the circle’s center, analogous to a bicycle traveling in a circle. To avoid commanding an instantaneous acceleration at the start of the maneuver, the trajectory includes a 10-second ramp generated using a quintic "smootherstep" interpolation function~\cite{quilez_smoothstep}. This polynomial ramp provides zero acceleration at both the start and end of the transition, ensuring that the payload smoothly accelerates from rest before entering the steady pirouette motion. A visualization of this trajectory is shown in Figure~\ref{fig:pirouette_3D_trajectory}.

Since the estimator is evaluated independently of the controller, the expected trajectory accelerations (i.e., feed-forward accelerations) are provided as inputs to the estimator, rather than the total payload accelerations produced by the controller, as described in Eq.~\ref{system:input}. Using the total payload accelerations would couple the estimator’s behavior to the controller’s tracking performance, making it difficult to attribute estimation errors to the estimator itself.

\subsubsection{Full rate communication} \label{s:piroutte_comms}
\begin{figure}
  \centering  
    \includegraphics[width=6in]{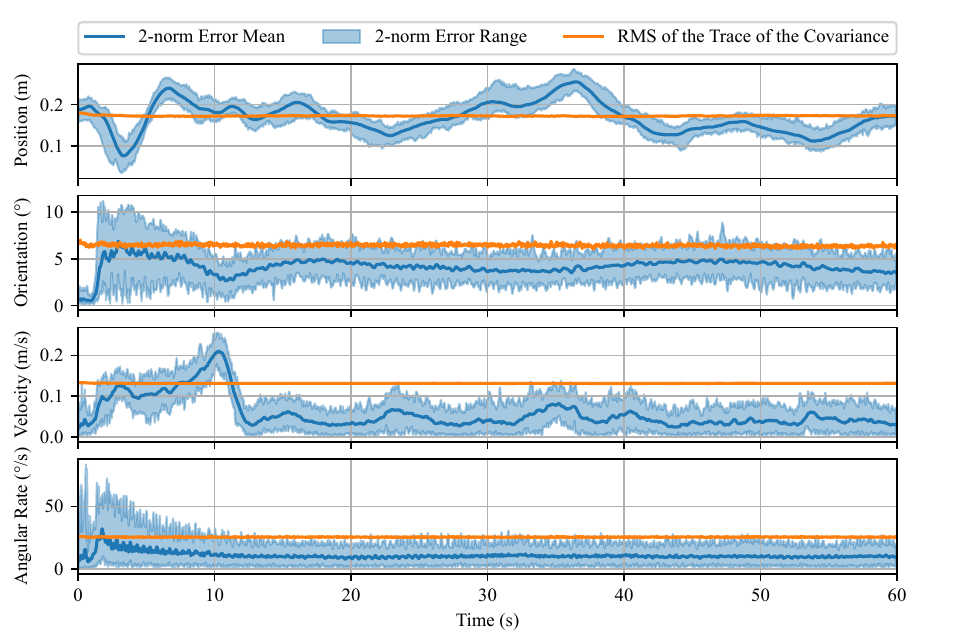}
  \caption{Pirouetting circle state estimate error.} 
  \label{fig:piroutte_state_error}
\end{figure}

First, Monte Carlo simulation results of the nominal case are presented where all rotorcraft maintain perfect communication, enabling the individual state estimators to share and fuse information from all four camera measurements. To summarize the results, at each timestep, $k$, across all runs, we report the minimum, maximum, and mean 2-norm error between the payload’s ground truth and its state estimate, as well as the root-mean-squared (RMS) value of the trace of the covariance matrix, $\mathbf{P}$, as shown in Figure~\ref{fig:piroutte_state_error}.

The close agreement between the mean error magnitude and the filter covariance indicates consistent and well-tuned estimator performance across all states. Larger peaks in the 2-norm error are observed during the initial portion of the trajectory (within the first 10 seconds), corresponding to the "smootherstep" ramp described previously. During this phase, the payload transitions from rest into the pirouette motion, which introduces minor velocity error and angular rate oscillations as the estimator adapts to the increasing acceleration profile. Once the trajectory ramp is complete, these effects diminish and the estimator settles into its steady-state performance.

\begin{figure}
  \centering
    \includegraphics[width=4in]{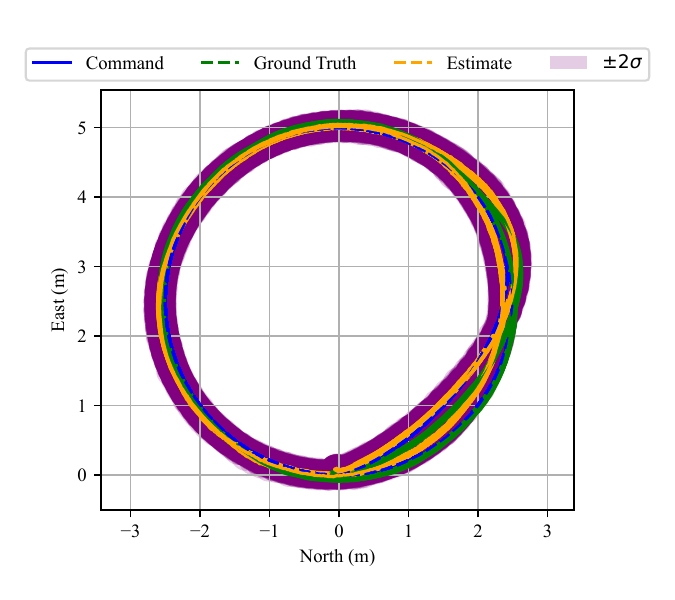}
  \caption{Pirouetting circle North-East position tracking projection.} 
  \label{fig:piroutte_xy_trajectory}
\end{figure}

To further illustrate the performance of the state estimator, Figure~\ref{fig:piroutte_xy_trajectory} shows a two-dimensional projection of the commanded, true, and estimated payload trajectories in the North-East plane, along with the associated uncertainty. The runs of the Monte Carlo simulation are overlaid to highlight the repeatability of the estimator across trials. The uncertainty bounds correspond to $\pm 2\sigma$, which represent the $95\%$ confidence region implied by the estimator’s covariance. Including these bounds allows for a visual assessment of estimator consistency.

\subsubsection{Loss of communication}
Next, we consider a scenario in which hardware faults, environmental effects, or tactical interference cause the rotorcraft to lose communication with one another. To represent this in the Monte Carlo simulations, all inter-vehicle communication was severed 20 seconds into each run and fully restored at 40 seconds. During this period, each drone relies solely on its own AprilTag measurement; thus, in Equations~\ref{information:y_update} and~\ref{information:Y_update}, no information from other drones is included ($N = 1$). The same result summarization method used in the full-communication case, namely, the 2-norm error and RMS of the trace of the covariance was applied.
\begin{figure}
  \centering  
    \includegraphics[width=6in]{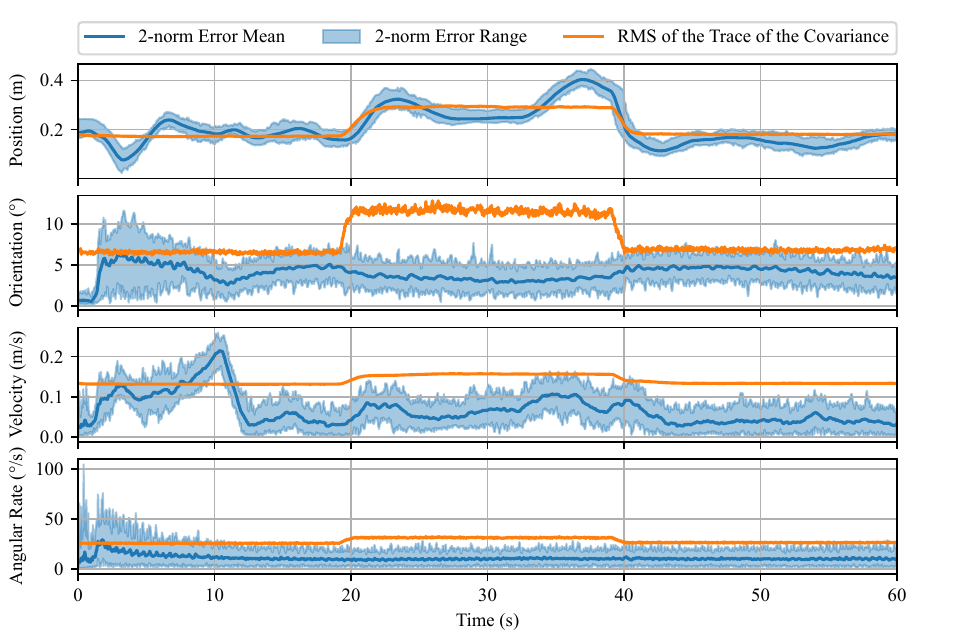}
  \caption{Pirouetting circle state estimate error under temporary communication loss.} 
  \label{fig:piroutte_state_error_commsLoss}
\end{figure}

Figure~\ref{fig:piroutte_state_error_commsLoss} shows that when communication between rotorcraft is lost near 20 seconds, the proposed state estimator continues to maintain a valid state estimate, with an increase in uncertainty that is expected given the loss of information. Instead of fusing measurements from four vehicles, the estimator must temporarily rely on only a single local measurement source, naturally leading to a larger covariance. While the loss of communication has little effect on the orientation, velocity, and angular-rate estimates, it results in a noticeably larger position 2-norm error. In contrast, a traditional batched Extended Kalman Filter would be unable to operate in this scenario, as it requires access to all measurements from every vehicle at every instant. The distributed information filter formulation therefore provides a key advantage: each drone can continue estimating independently during communication outages and seamlessly reintegrate shared information once connectivity is restored.

\begin{figure}
  \centering  
    \includegraphics[width=4in]{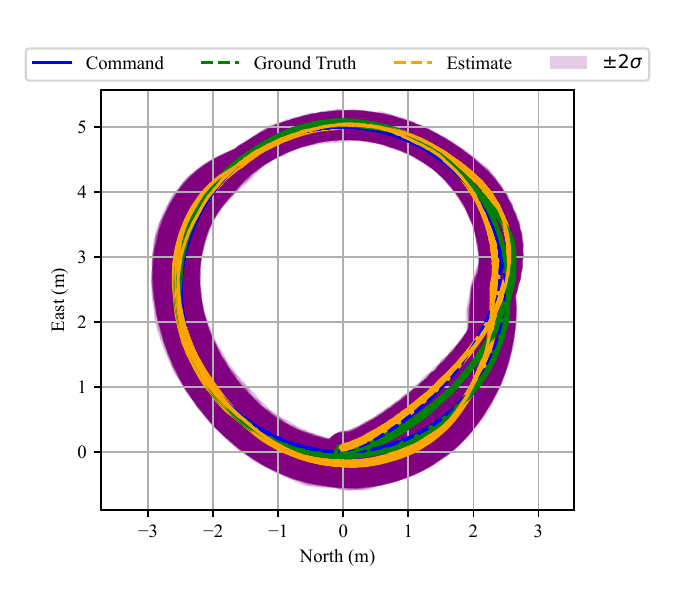}
  \caption{Pirouetting circle position tracking North-East projection under temporary communication loss.} 
  \label{fig:piroutte_xy_trajectory_commsLoss}
\end{figure}

To help illustrate the change in uncertainty, Figure~\ref{fig:piroutte_xy_trajectory_commsLoss} shows a two-dimensional projection of the commanded, true, and estimated payload trajectories in the North-East plane, along with the associated uncertainty. An expanded uncertainty band becomes evident near -2 meters North and 4 meters East, growing counter-clockwise as the temporary communication loss begins. Once communication is restored near 2 meters North and 2 meters East, the uncertainty band contracts back to its original width, reflecting the return of additional measurement information.

\subsection{Estimator in the loop}\label{s:estimator_intheloop}
After validating the proposed extended information filter independently from the multilift system, it is next integrated directly into the multilift control loop as the primary source of state feedback, replacing the Gazebo ground truth data used in the previous section. This configuration enables evaluation of the closed-loop performance of the multilift system under realistic sensor noise and disturbances, demonstrating the practical effectiveness of the proposed estimator within the full control architecture.

\begin{figure}
  \centering  
    \includegraphics[width=3.5in]{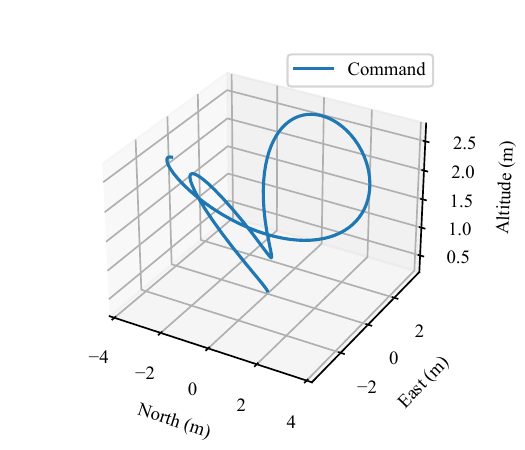}
  \caption{Commanded Lissajous figure-8 trajectory.} 
  \label{fig:lissajous_3D_trajectory}
\end{figure}

\begin{figure}
    \centering
    
    \begin{subfigure}{0.32\textwidth}
        \centering
        \includegraphics[width=\linewidth]{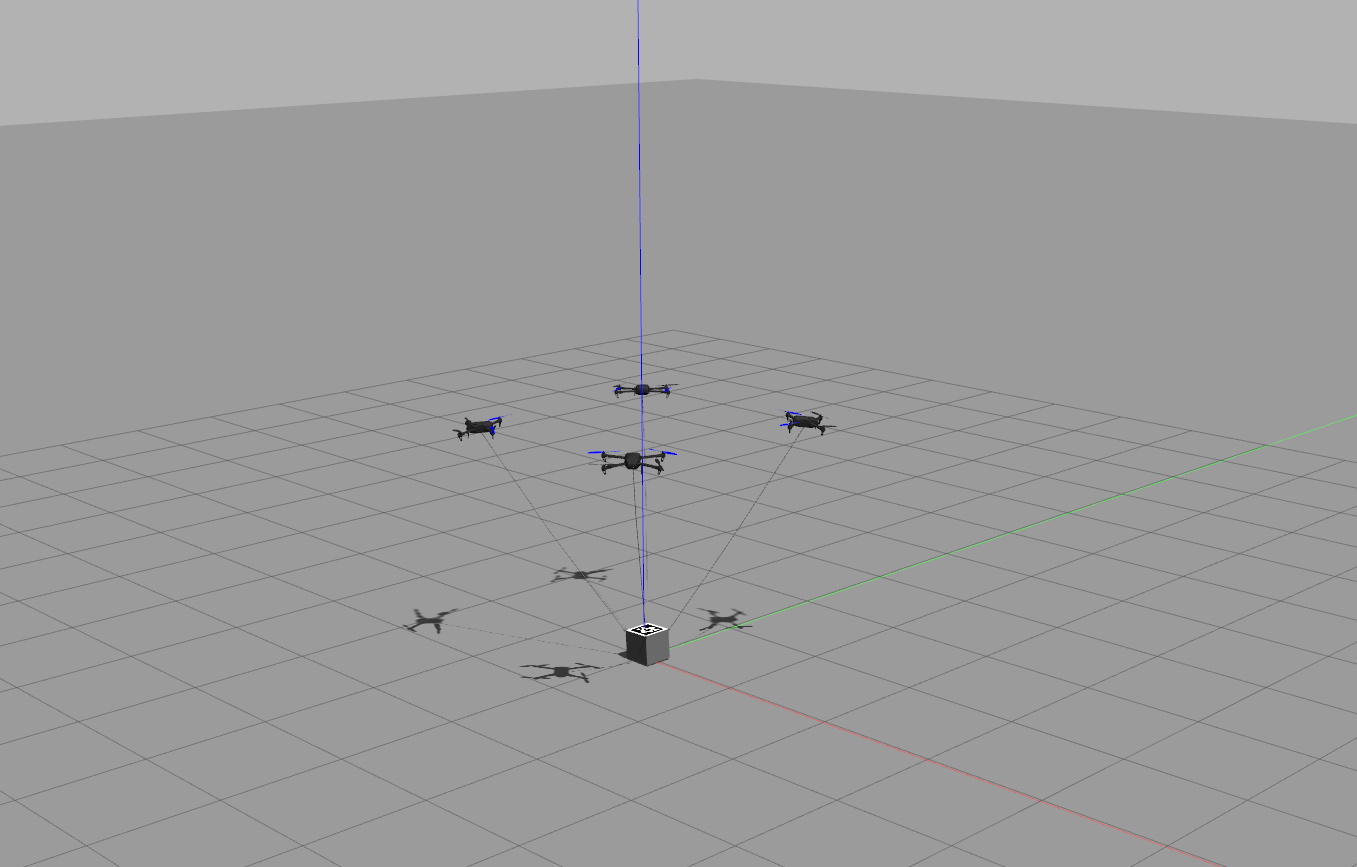}
        \caption{Time = 0s}
    \end{subfigure}
    \begin{subfigure}{0.32\textwidth}
        \centering
        \includegraphics[width=\linewidth]{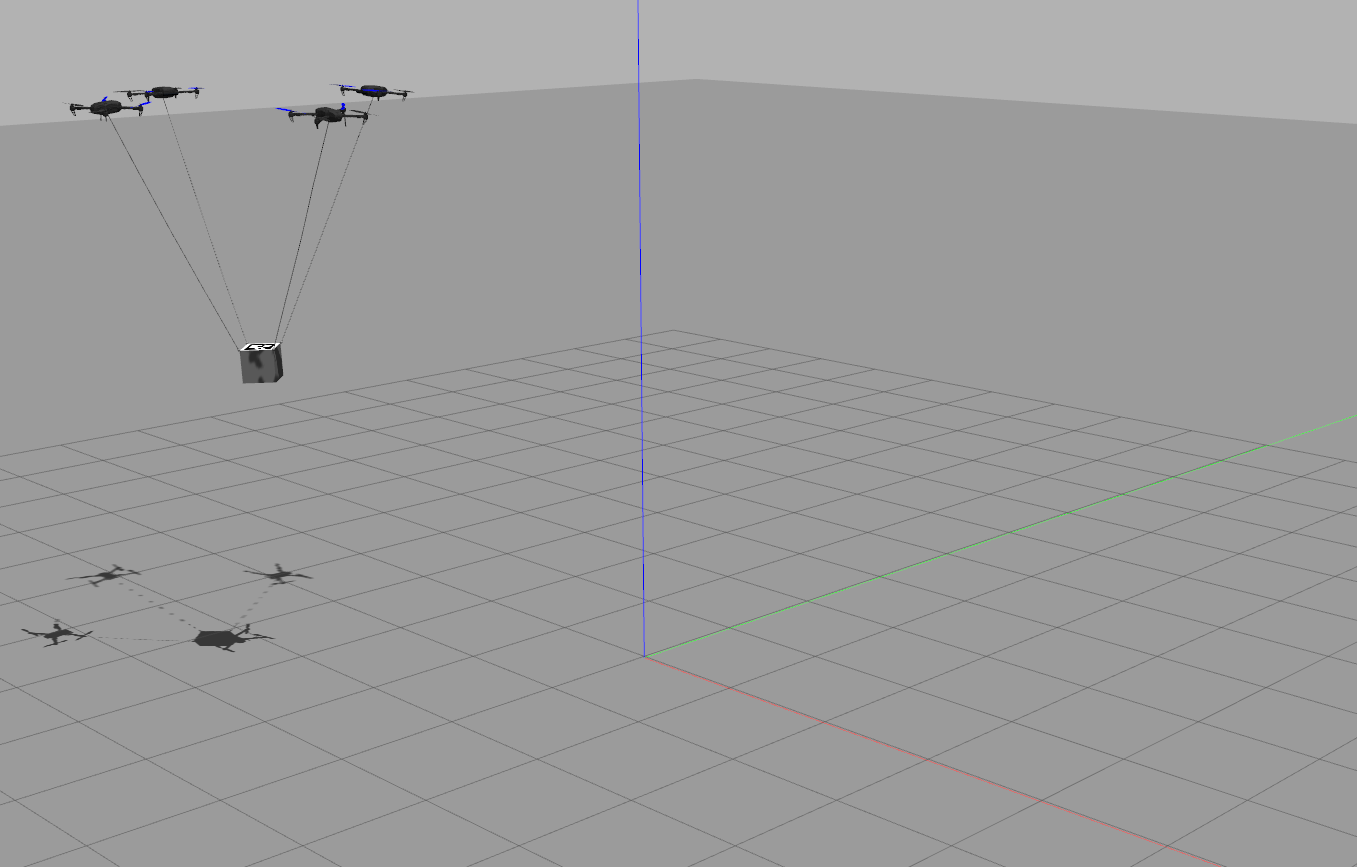}
        \caption{Time = 10s}
    \end{subfigure}
    \begin{subfigure}{0.32\textwidth}
        \centering
        \includegraphics[width=\linewidth]{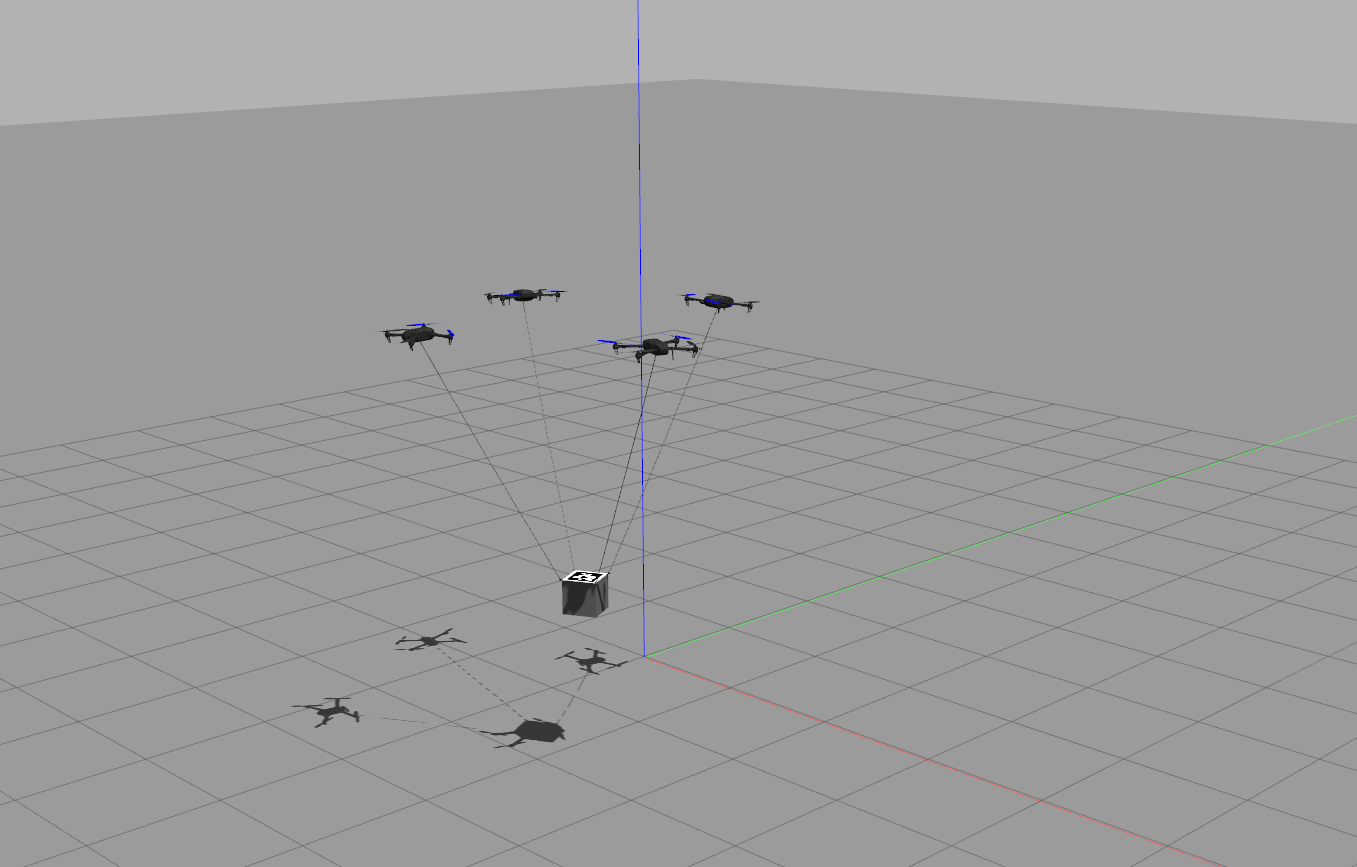}
        \caption{Time = 20s}
    \end{subfigure}

    \begin{subfigure}{0.32\textwidth}
        \centering
        \includegraphics[width=\linewidth]{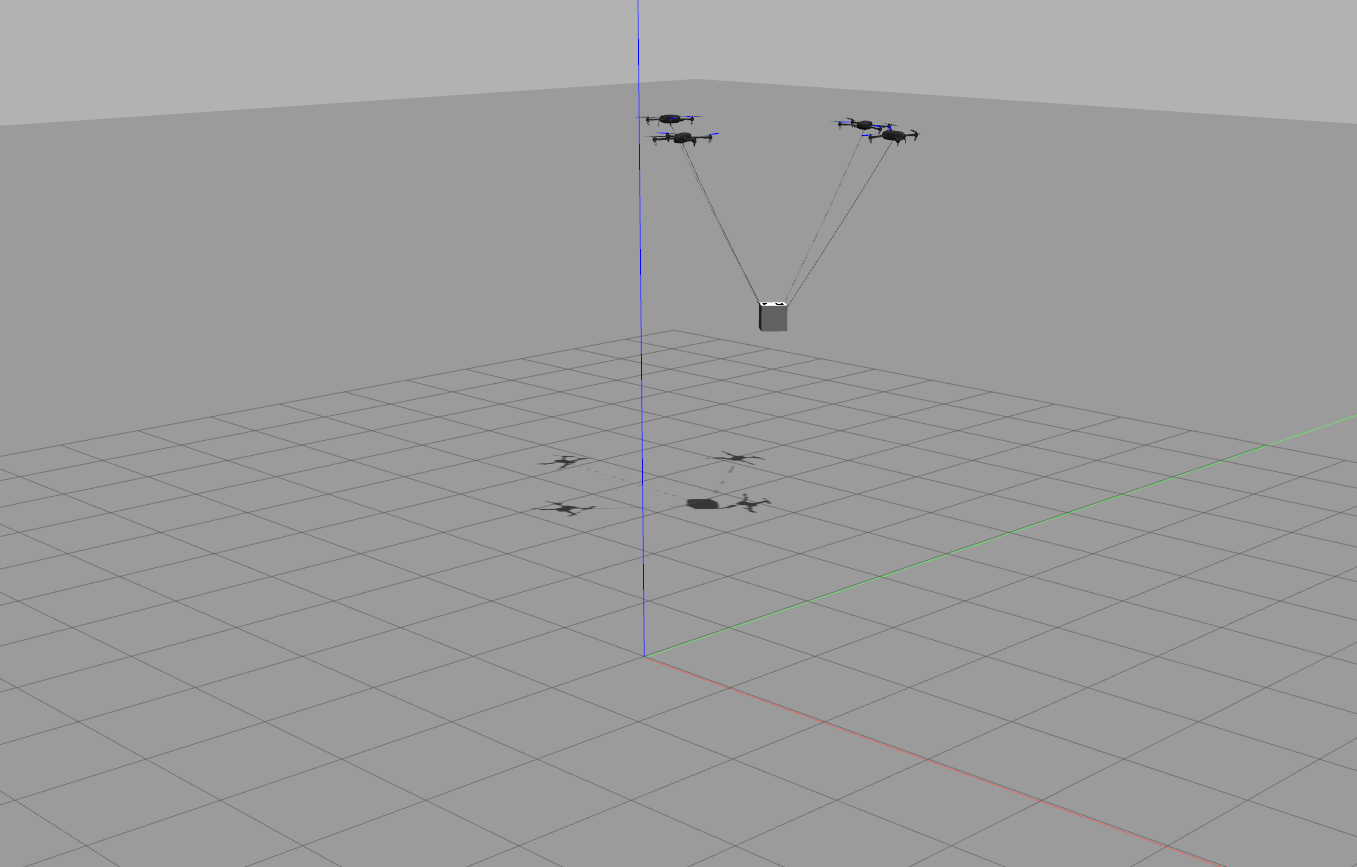}
        \caption{Time = 30s}
    \end{subfigure}
    \begin{subfigure}{0.32\textwidth}
        \centering
        \includegraphics[width=\linewidth]{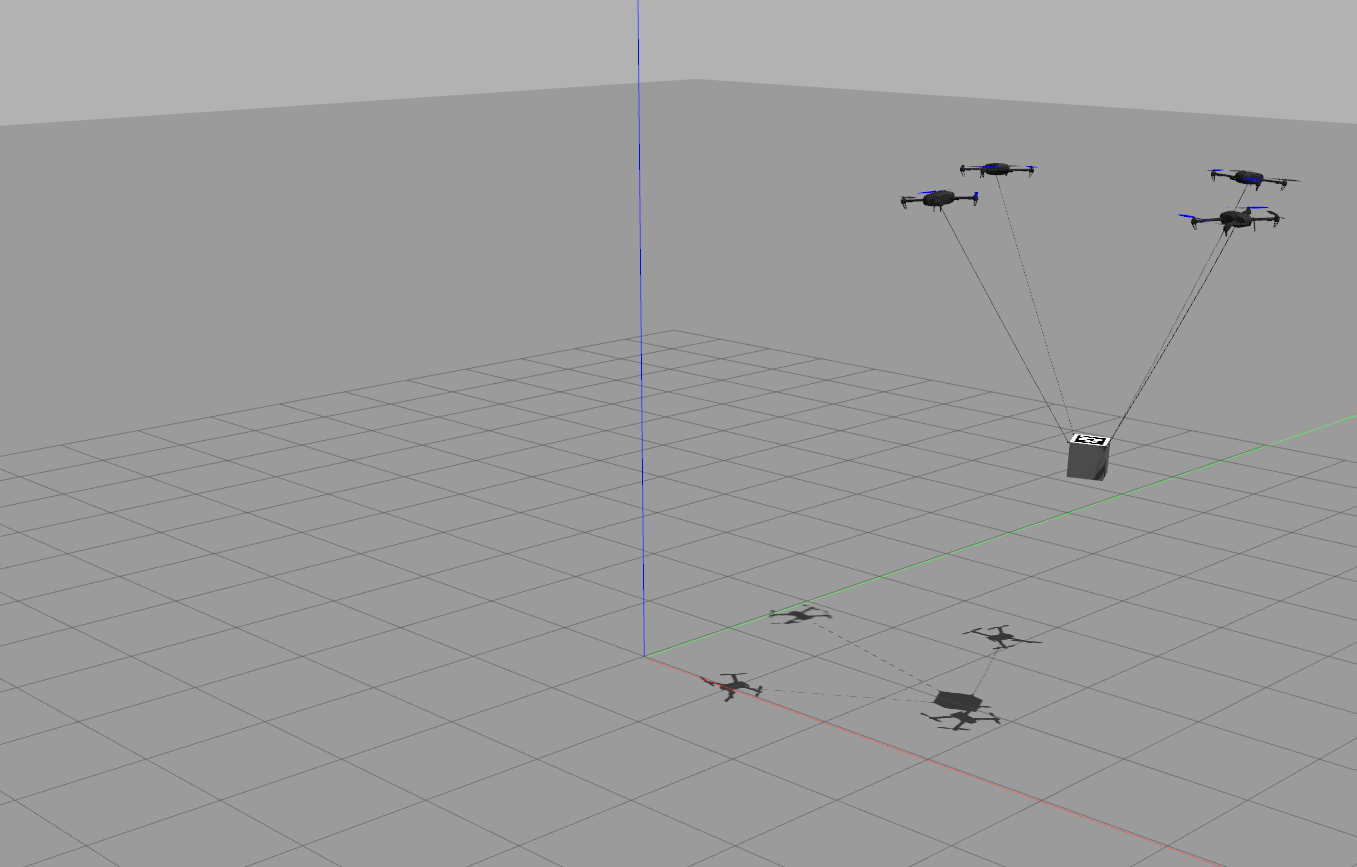}
        \caption{Time = 40s}
    \end{subfigure}
    \begin{subfigure}{0.32\textwidth}
        \centering
        \includegraphics[width=\linewidth]{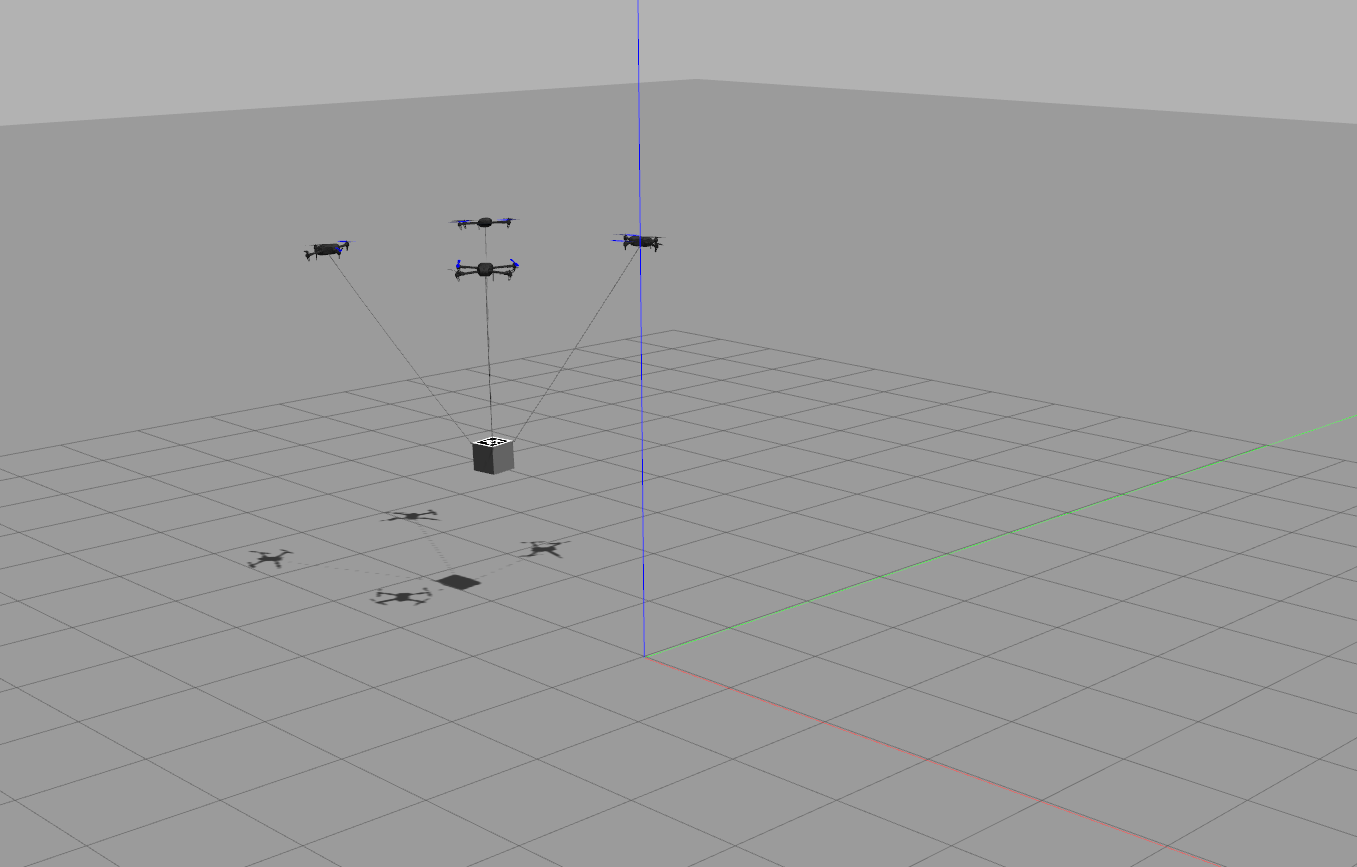}
        \caption{Time = 50s}
    \end{subfigure}

    \caption{The multilift Gazebo system flying the commanded Lissajous figure-8 trajectory.}
    \label{fig:lissajous_gazebo_trajectory}
\end{figure}

For the following results, a figure-8 Lissajous-style trajectory was used to challenge the estimator with coupled, time-varying motion while maintaining smooth, continuous position, velocity, and acceleration profiles. The horizontal motion was generated with sinusoidal components $f_n=0.04$ and $f_e=0.02$ Hz, while the vertical motion includes a small oscillation tied to the same frequency as $f_n$. As with the pirouette trajectory discussed earlier, a "smootherstep" ramp over 15 seconds was applied to ensure a gradual increase in motion and to avoid abrupt accelerations at the start. Throughout the trajectory, the payload is oriented to continually face the direction of travel, analogous to a bicycle navigating a curving path. A visualization of the trajectory is shown in Figure~\ref{fig:lissajous_3D_trajectory} and Figure~\ref{fig:lissajous_gazebo_trajectory}.

\subsubsection{Full rate communication}
\begin{figure}
  \centering  
    \includegraphics[width=6in]{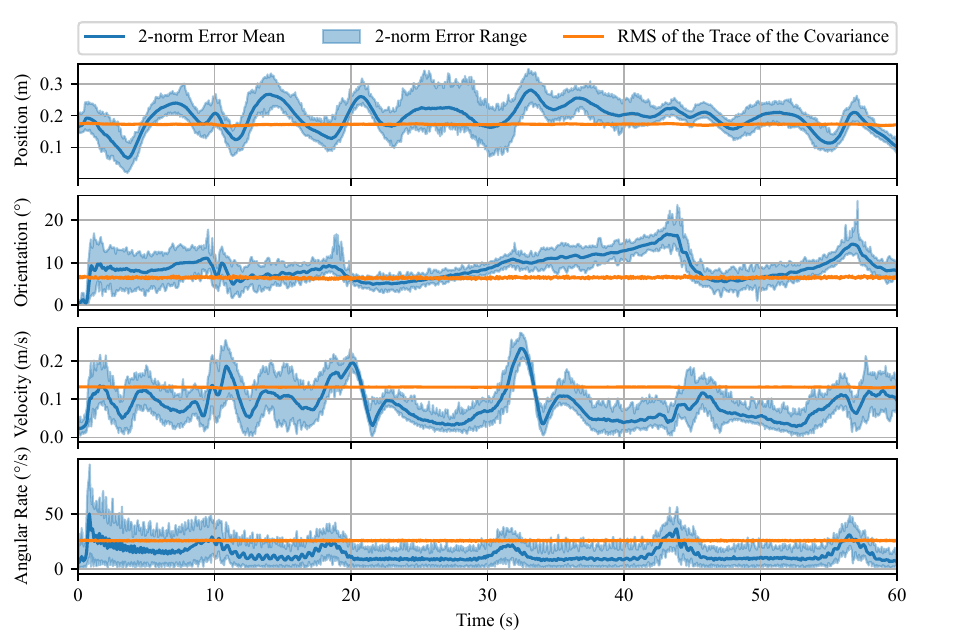}
  \caption{Lissajous figure-8 trajectory state estimate error.} 
  \label{fig:lissajous_state_error}
\end{figure}

As in the pirouette case, Figure~\ref{fig:lissajous_state_error} reports the minimum, maximum, and mean 2-norm state estimation error across all Monte Carlo runs, along with the RMS trace of the covariance matrix. The general interpretation of this plot follows the description provided in Section~\ref{s:piroutte_comms}; here we focus on the differences introduced by the Lissajous motion. As shown in the angular rate subplot, the peaks correspond to the corners of the Lissajous trajectory where large yaw rate commands occur. At these time instances, the 2-norm errors in orientation and velocity exhibit corresponding peaks that rise above the estimator’s average uncertainty level. This behavior is expected as at the trajectory’s large yaw rate momentarily amplifies the effects of process and measurement noise. These brief periods of aggressive motion therefore produce larger error than during steady segments of the trajectory, leading to the observed error peaks.

\begin{figure}
  \centering  
    \includegraphics[width=6in]{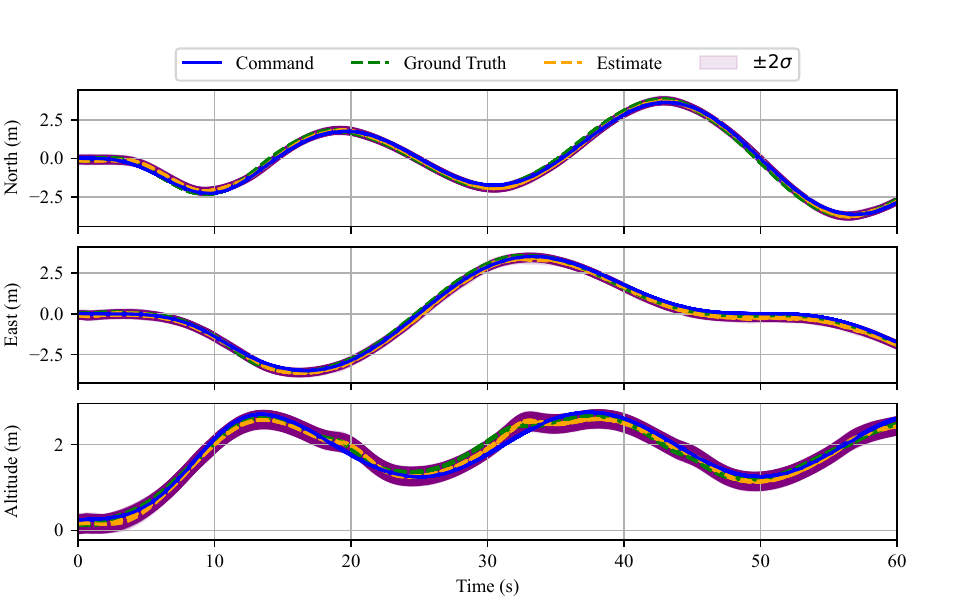}
  \caption{Lissajous trajectory information filter in-the-loop position tracking performance.}
  \label{fig:lissajous_position_uncertainty}
\end{figure}

\begin{figure}
  \centering  
    \includegraphics[width=6in]{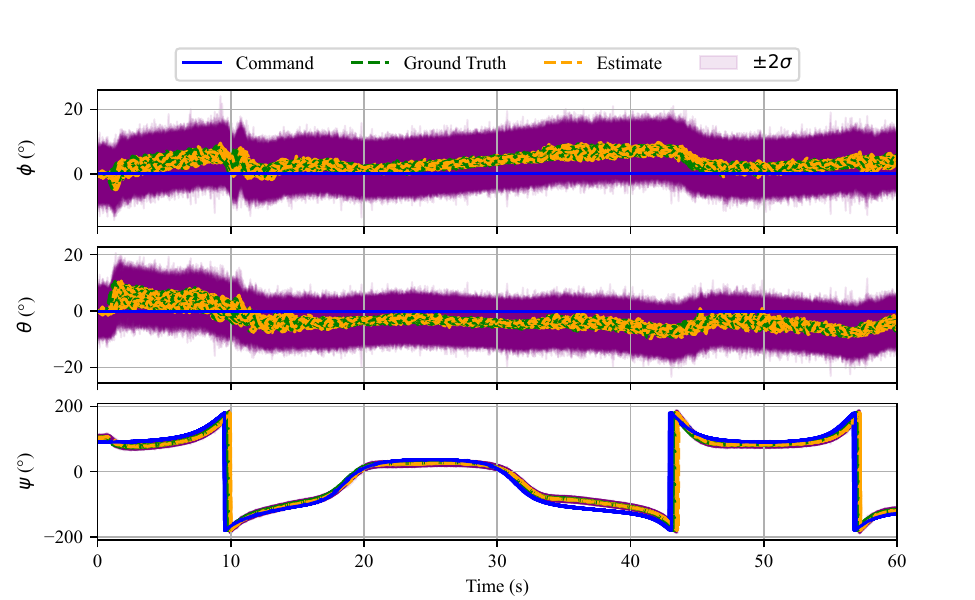}
  \caption{Lissajous trajectory information filter in-the-loop orientation tracking performance.}
  \label{fig:lissajous_orientation_uncertainty}
\end{figure}

To illustrate the interaction between the estimator and the multilift controller, Figure~\ref{fig:lissajous_position_uncertainty} and Figure~\ref{fig:lissajous_orientation_uncertainty} show the commanded, true, and estimated payload trajectories for position and orientation respectively, along with their corresponding $\pm 2\sigma$ uncertainty bounds. Because the controller uses the state estimate as its feedback block, accurate estimation is required for good tracking performance. The results show that the controller follows the commanded trajectory closely and that the estimator remains consistent, with errors staying well within the predicted covariance throughout the trajectory.

\subsubsection{Loss of communication}
\begin{figure}
  \centering  
    \includegraphics[width=6in]{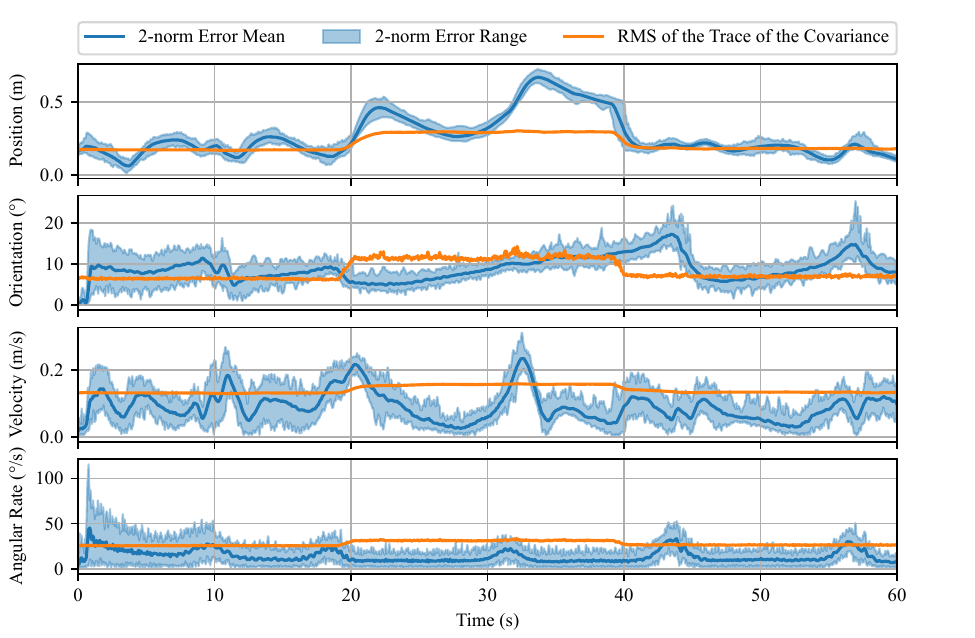}
  \caption{Lissajous figure-8 trajectory state estimate error under a period of communication loss.} 
  \label{fig:lissajous_state_error_commsLoss}
\end{figure}

A similar communication loss scenario was evaluated for the Lissajous figure-8 trajectory, now with the estimator operating directly inside the multilift control loop. As in the pirouette case, all inter-vehicle communication is lost at 20 seconds and restored at 40 seconds, during which each drone relies solely on its own AprilTag measurements. The same error metrics (2-norm state error and RMS trace of the covariance) are used to summarize performance. Figure~\ref{fig:lissajous_state_error_commsLoss} shows that the loss of communication leads to a predictable increase in estimator uncertainty, just as observed in the pirouette scenario. Larger 2-norm errors are particularly noticeable in the position estimates during the communication dropout compared to the full communication scenario.

\begin{figure}
  \centering  
    \includegraphics[width=6in]{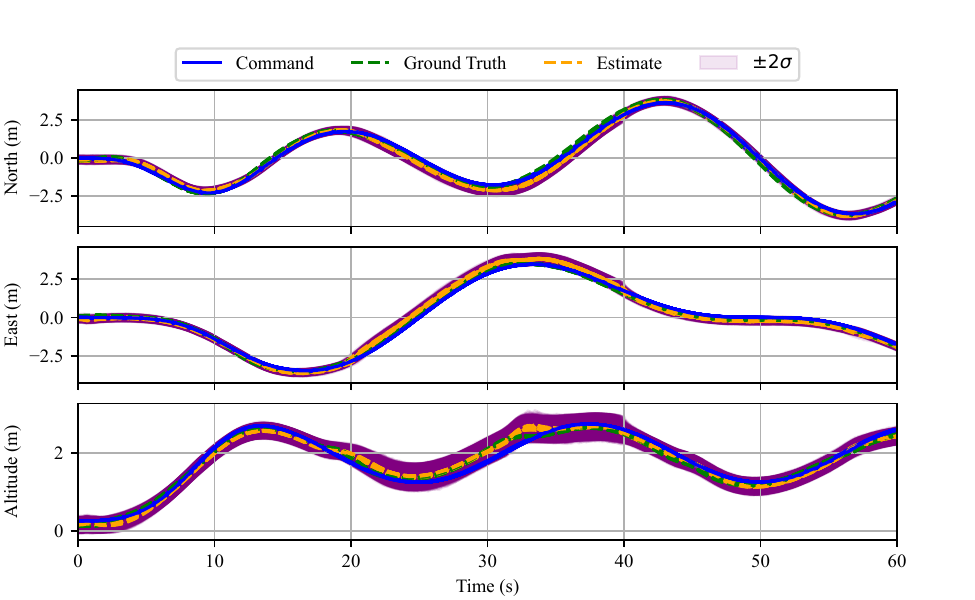}
  \caption{Lissajous trajectory information filter in-the-loop position tracking performance during temporary communication loss.}
  \label{fig:lissajous_position_uncertainty_commsLoss}
\end{figure}

\begin{figure}
  \centering  
    \includegraphics[width=6in]{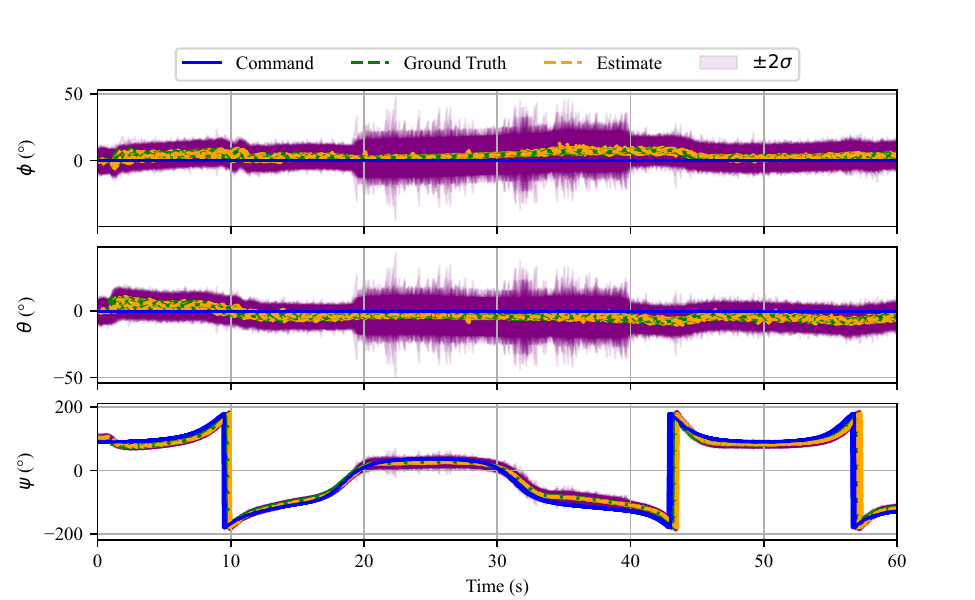}
  \caption{Lissajous trajectory information filter in-the-loop orientation tracking performance during temporary communication loss.}
  \label{fig:lissajous_orientation_uncertainty_commsLoss}
\end{figure}

To provide a more detailed view of how the estimator interacts with the controller, Figure~\ref{fig:lissajous_position_uncertainty_commsLoss} and Figure~\ref{fig:lissajous_orientation_uncertainty_commsLoss} present the payload’s position and orientation trajectories, respectively, along with the corresponding $\pm 2\sigma$ uncertainty bounds. These plots show that even during the communication outage, the controller maintains close tracking of the commanded trajectory, though performance is slightly degraded when relying on the state estimate without shared information. Once connectivity is restored, the covariance rapidly contracts, and both position and orientation tracking return to nominal performance, demonstrating the combined robustness of the estimator and multilift control loop under temporary communication loss.

\section{Conclusion} \label{s:conclusion}
This work presented a distributed approach to payload state estimation for a cooperative multilift system, offering a more controllable and economic solution for transporting "long tail" payloads. By replacing the onboard sensor suite mounted on the payload with an AprilTag fiducial marker, the proposed method enables operation in GPS-denied environments, enhances system reliability, and reduces deployment complexity.

A distributed information filter framework was implemented in a software in the loop simulation using Gazebo. The results demonstrate that the team of rotorcraft can compute an accurate payload state estimate when sharing information, and can maintain this estimate even during periods of communication loss. Moreover, when operating within the multilift control loop, the estimator provided consistent and accurate feedback, allowing the controller to maintain stable and precise trajectory tracking. These findings highlight both the robustness and practical applicability of the approach in realistic operational scenarios.

\section*{Acknowledgments}
This research was funded by the Naval Air Warfare Center Aircraft Division under Grant N00421-24-1-0001. The views and conclusions contained in this document are those of the authors and should not be interpreted as representing the official policies or position, either expressed or implied, of the Naval Air Warfare Center Aircraft Division (NAWCAD) or the U.S. Government.

\bibliography{estimation}

@inproceedings{BerKon09,
    title={Generic slung load transportation system using small size helicopters},
    author={M. Bernard and K. Kondak},
    booktitle={{IEEE} International Conference on Robotics and Automation},
    pages={3258--3264},
    year={2009},
    organization={IEEE}
}

@techreport{CicKan92,
  author= {L. S. Cicolani and G. Kanning},
  title= {Equations of Motion of Slung-Load Systems, Including Multilift Systems},
  institution = {NASA},
  number={TM-1038798},
  year= {1992}
}

@inproceedings{GenLan19,
    title={Implementation and demonstration of coordinated transport of a slung load by a team of rotorcraft},
    author={J. Geng and J. W. Langelaan},
    booktitle={AIAA SciTech 2019 Forum},
    year={2019},
}

@inproceedings{LiHor14,
    title={Coordinated Transport of a Slung Load by a Team of Autonomous Rotorcraft},
    author={Z. Li and J. F. Horn and J. W. Langelaan},
    booktitle={AIAA Guidance, Navigation, and Control Conference},
    year={2014},
}

@techreport{MeeChe70,
  author= {T. Meek and G. Chesley, et al.},
  title= {Twin helicopter lift system study and feasibility demonstration},
  institution = {Sikorsky Engineering Report},
  number={64323},
  year= {1970}
}

@article{MicFin11,
  title={Cooperative manipulation and transportation with aerial robots},
  author={N. Michael and J. Fink and V. Kumar},
  journal={Autonomous Robots},
  volume={30},
  number={1},
  year={2011},
  pages={73--86}
}

@article{MitPra91,
  title={Nonlinear adaptive control of a twin lift helicopter system},
  author={M. Mittal and J. Prasad and D. P. Schrage},
  journal={IEEE Control Systems},
  volume={11},
  number={2},
  year={1991},
  pages={39--45}
}

@article{MitPra92,
  title={Comparison of stability and control characteristics of two twin-lift helicopter configurations},
  author={M. Mittal and J. Prasad and D. P. Schrage},
  journal={Nonlinear Dynamics},
  volume={3},
  number={3},
  year={1992},
  pages={199-223}
}

@article{MitPra93,
  title={Three-dimensional modeling and control of a twin-lift helicopter system},
  author={M. Mittal and J. Prasad},
  journal={Journal of Guidance, Control and Dynamics},
  volume={16},
  number={1},
  year={1993},
  pages={86--95}
}

@inproceedings{hendrick2022scaled,
  title={Scaled Experiments in Flight Control Design for Autonomous Landing in High Sea States},
  author={Hendrick, Christopher M and Nicholson, Duncan and Jaques, Emma R and Horn, Joseph and Langelaan, Jack W and Sydney, Anish J},
  booktitle={AIAA AVIATION 2022 Forum},
  pages={3280},
  year={2022}
}

@inproceedings{nicholson2022scaled,
  title={Scaled Experiments in Vision-Based Approach and Landing in High Sea States},
  author={Nicholson, Duncan and Hendrick, Christopher M and Jaques, Emma R and Horn, Joseph and Langelaan, Jack W and Sydney, Anish J},
  booktitle={AIAA AVIATION 2022 Forum},
  pages={3279},
  year={2022}
}

@article{liu2022robotic,
  title={Robotic Depowdering for Additive Manufacturing Via Pose Tracking},
  author={Liu, Zhenwei and Geng, Junyi and Dai, Xikai and Swierzewski, Tomasz and Shimada, Kenji},
  journal={IEEE Robotics and Automation Letters},
  volume={7},
  number={4},
  pages={10770--10777},
  year={2022},
  publisher={IEEE}
}

@article{geng2020cooperative,
  title={Cooperative transport of a slung load using load-leading control},
  author={Geng, Junyi and Langelaan, Jack W},
  journal={Journal of Guidance, Control, and Dynamics},
  volume={43},
  number={7},
  pages={1313--1331},
  year={2020},
  publisher={American Institute of Aeronautics and Astronautics}
}

@article{geng2021estimation,
  title={Estimation of inertial properties for a multilift slung load},
  author={Geng, Junyi and Langelaan, Jack W},
  journal={Journal of Guidance, Control, and Dynamics},
  volume={44},
  number={2},
  pages={220--237},
  year={2021},
  publisher={American Institute of Aeronautics and Astronautics}
}

@article{geng2022load,
  title={Load-Distribution-Based Trajectory Planning and Control for a Multilift System},
  author={Geng, Junyi and Singla, Puneet and Langelaan, Jack W},
  journal={Journal of Aerospace Information Systems},
  volume={19},
  number={5},
  pages={366--381},
  year={2022},
  publisher={American Institute of Aeronautics and Astronautics}
}

@article{deng2022icaps,
  title={iCaps: Iterative Category-Level Object Pose and Shape Estimation},
  author={Deng, Xinke and Geng, Junyi and Bretl, Timothy and Xiang, Yu and Fox, Dieter},
  journal={IEEE Robotics and Automation Letters},
  volume={7},
  number={2},
  pages={1784--1791},
  year={2022},
  publisher={IEEE}
}

@article{hu2023off,
  title = {Off-Policy Evaluation With Online Adaptation for Robot Exploration in Challenging Environments},
  author = {Hu, Yafei and Geng, Junyi and Wang, Chen and Keller, John and Scherer, Sebastian},
  journal = {IEEE Robotics and Automation Letters (RA-L)},
  volume = {8},
  number = {6},
  pages = {3780-3787},
  year = {2023},
  publisher = {IEEE}
}

@inproceedings{zhao2021super,
  title={Super odometry: IMU-centric LiDAR-visual-inertial estimator for challenging environments},
  author={Zhao, Shibo and Zhang, Hengrui and Wang, Peng and Nogueira, Lucas and Scherer, Sebastian},
  booktitle={2021 IEEE/RSJ International Conference on Intelligent Robots and Systems (IROS)},
  pages={8729--8736},
  year={2021},
  organization={IEEE}
}

@article{qin2018vins,
  title={Vins-mono: A robust and versatile monocular visual-inertial state estimator},
  author={Qin, Tong and Li, Peiliang and Shen, Shaojie},
  journal={IEEE Transactions on Robotics},
  volume={34},
  number={4},
  pages={1004--1020},
  year={2018},
  publisher={IEEE}
}

@article{li2021cooperative,
  title={Cooperative transportation of cable suspended payloads with mavs using monocular vision and inertial sensing},
  author={Li, Guanrui and Ge, Rundong and Loianno, Giuseppe},
  journal={IEEE Robotics and Automation Letters},
  volume={6},
  number={3},
  pages={5316--5323},
  year={2021},
  publisher={IEEE}
}

@article{sanalitro2020full,
  title={Full-pose manipulation control of a cable-suspended load with multiple UAVs under uncertainties},
  author={Sanalitro, Dario and Savino, Heitor J and Tognon, Marco and Cort{\'e}s, Juan and Franchi, Antonio},
  journal={IEEE Robotics and Automation Letters},
  volume={5},
  number={2},
  pages={2185--2191},
  year={2020},
  publisher={IEEE}
}

@article{zhang2021self,
  title={Self-triggered based coordinate control with low communication for tethered multi-UAV collaborative transportation},
  author={Zhang, Xiaozhen and Zhang, Fan and Huang, Panfeng and Gao, Jiale and Yu, Hang and Pei, Chongxu and Zhang, Yizhai},
  journal={IEEE Robotics and Automation Letters},
  volume={6},
  number={2},
  pages={1559--1566},
  year={2021},
  publisher={IEEE}
}

@article{superpumafirefighting,
	Author = {Jackson, Paul},
	Journal = {Air International},
	Number = {1},
	Pages = {7-12, 33-35},
	Title = {Super Puma},
	Volume = {26},
	Year = {1984}}

@article{hutto1976flight,
  title={Flight-Test Report on the Heavy-Lift Helicopter Flight-Control System},
  author={Hutto, AJ},
  journal={Journal of the American Helicopter Society},
  volume={21},
  number={1},
  pages={32--40},
  year={1976},
  publisher={Vertical Flight Society}
}

@inproceedings{murray1996trajectory,
	Author = {Murray, Richard M},
	Booktitle = {IFAC Proceedings Volumes},
	Title = {Trajectory generation for a towed cable system using differential flatness},
	Volume={29},
    Number={1},
    Pages={2792-2797},
	Year = {1996},
	Publisher={Elsevier}
}

@article{smirnov1990multiple,
  title={Multiple-power-path nonplanetary main gearbox of the Mi-26 heavy-lift transport helicopter},
  author={SMIRNOV, GENNADII},
  journal={Vertiflite},
  volume={36},
  pages={20--23},
  year={1990}
}

@inproceedings{carter1982implication,
  address = {Aix-en-Provence, France},
  title={Implication of heavy lift helicopter size effect trends and multilift options for filling the need},
  author={Carter, Edward S},
  booktitle={European Rotorcraft Forum, 8 th},
  year={1982}
}

@article{horn2023experimental,
  title={Experimental Analysis of Advanced Control and Estimation Systems for Autonomous Ship Landing},
  author={Horn, Joseph F and Langelaan, Jack W},
  year={2023}
}

@book{Mutambara1998,
    author = {Arthur G. O. Mutambara},
    title = {Decentralized Estimation and Control for Multisensor Systems},
    publisher = {CRC Press},
    year = {1998}
}

@inproceedings{Lee08,
    author = {Deok-Jin Lee},
    title = {Unscented Information Filtering for Distributed Estimation and Multiple Sensor Fusion},
    booktitle = {AIAA Guidance, Navigation and Control Conference},
    year = {2008},
    month = {August}
}

@INPROCEEDINGS{OlsonAprilTag2011,
  author={Olson, Edwin},
  booktitle={2011 IEEE International Conference on Robotics and Automation}, 
  title={AprilTag: A robust and flexible visual fiducial system}, 
  year={2011},
  volume={},
  number={},
  pages={3400-3407},
  doi={10.1109/ICRA.2011.5979561}}

@article{Kalaitzakis2021,
  author = {Kalaitzakis, Michail and Cain, Brennan and Carroll, Sabrina and Ambrosi, Anand and Whitehead, Camden and Vitzilaios, Nikolaos},
  title = {Fiducial Markers for Pose Estimation},
  journal = {Journal of Intelligent \& Robotic Systems},
  volume = {101},
  number = {4},
  pages = {71},
  year = {2021},
  doi = {10.1007/s10846-020-01307-9},
  url = {https://doi.org/10.1007/s10846-020-01307-9}
}

@misc{quilez_smoothstep,
  author = {Quilez, Inigo},
  title = {Smoothstep and Smootherstep Interpolation Functions},
  url = {https://iquilezles.org/articles/smoothsteps/},
  year = {2008}
}

\end{document}